\PassOptionsToPackage{prologue,table}{xcolor}
\documentclass[sigconf]{acmart}
\AtBeginDocument{%
  }
\copyrightyear{2025}
\acmYear{2025}
\setcopyright{acmlicensed}\acmConference[MM '25]{Proceedings of the 33rd ACM International Conference on Multimedia}{October 27--31, 2025}{Dublin, Ireland}
\acmBooktitle{Proceedings of the 33rd ACM International Conference on Multimedia (MM '25), October 27--31, 2025, Dublin, Ireland}
\acmDOI{10.1145/3746027.3754498}
\acmISBN{979-8-4007-2035-2/2025/10}
\usepackage{algorithm}
\usepackage{algorithmic}
\usepackage{bm}
\usepackage{enumitem}
\usepackage{xspace}
\usepackage{scalerel}
\usepackage{booktabs, caption, array}
\usepackage{multirow}
\newcommand{\highllm}{Pilot-LLM\xspace}

\newcommand{\lowuav}{MSP\xspace}
\begin{document}

\title{AeroDuo: Aerial Duo for UAV-based Vision and Language Navigation}

\author{Ruipu Wu}
\authornote{Equal contribution.}
\affiliation{
  \institution{Beihang University}
  \city{Beijing}
  \country{China}
}
\email{bastien_wu@buaa.edu.cn}

\author{Yige Zhang}
\authornotemark[1]
\affiliation{%
  \institution{Beihang University}
  \city{Beijing}
  \country{China}
}
\email{yige_zhang@buaa.edu.cn}

\author{Jinyu Chen}
\authornotemark[1]
\affiliation{%
  \institution{Beihang University}
  \city{Beijing}
  \country{China}
}
\email{chenjinyu@buaa.edu.cn}

\author{Linjiang Huang}
\authornote{Corresponding author.}
\affiliation{%
  \institution{Beihang University}
  \city{Beijing}
  \country{China}
}
\email{ljhuang@buaa.edu.cn}

\author{Shifeng Zhang}
\affiliation{%
  \institution{Sangfor Technologies Inc.}
  \city{Shenzhen}
  \country{China}
}
\email{zhangshifeng@sangfor.com.cn}

\author{Xu Zhou}
\affiliation{%
  \institution{Sangfor Technologies Inc.}
  \city{Shenzhen}
  \country{China}
}
\email{zhouxu@sangfor.com.cn}

\author{Liang Wang}
\affiliation{%
  \institution{Institute of Automation, Chinese Academy of Sciences}
  \city{Beijing}
  \country{China}
}
\email{wangliang@nlpr.ia.ac.cn}

\author{Si Liu}
\affiliation{%
  \institution{Beihang University}
  \city{Beijing}
  \country{China}
}
\email{liusi@buaa.edu.cn}

\renewcommand{\shortauthors}{Ruipu Wu et al.}

\begin{abstract}
Aerial Vision-and-Language Navigation (VLN) is an emerging task that enables Unmanned Aerial Vehicles (UAVs) to navigate outdoor environments using natural language instructions and visual cues. However, due to the extended trajectories and complex maneuverability of UAVs, achieving reliable UAV-VLN performance is challenging and often requires human intervention or overly detailed instructions.

To harness the advantages of UAVs’ high mobility, which could provide multi-grained perspectives, while maintaining a manageable motion space for learning, we introduce a novel task called Dual-Altitude UAV Collaborative VLN (DuAl-VLN). In this task, two UAVs operate at distinct altitudes: a high-altitude UAV responsible for broad environmental reasoning, and a low-altitude UAV tasked with precise navigation. To support the training and evaluation of the DuAl-VLN, we construct the HaL-13k, a dataset comprising $13,838$ collaborative high-low UAV demonstration trajectories, each paired with target-oriented language instructions. This dataset includes both unseen maps and an unseen object validation set to systematically evaluate the model's generalization capabilities across novel environments and unfamiliar targets.
To consolidate their complementary strengths, we propose a dual-UAV collaborative VLN framework, AeroDuo, where the high-altitude UAV integrates a multimodal large language model (Pilot-LLM) for target reasoning, while the low-altitude UAV employs a lightweight multi-stage policy for navigation and target grounding.
The two UAVs work collaboratively and only exchange minimal coordinate information to ensure efficiency.
Experimental results indicate that AeroDuo achieves an evident 9.71\% improvement in success rates compared to existing single-UAV methods, demonstrating the effectiveness of dual-altitude collaboration in balancing environmental coverage, precision, and operational autonomy.
\end{abstract}




\begin{teaserfigure}
    \centering
    \includegraphics[width=0.95\textwidth]{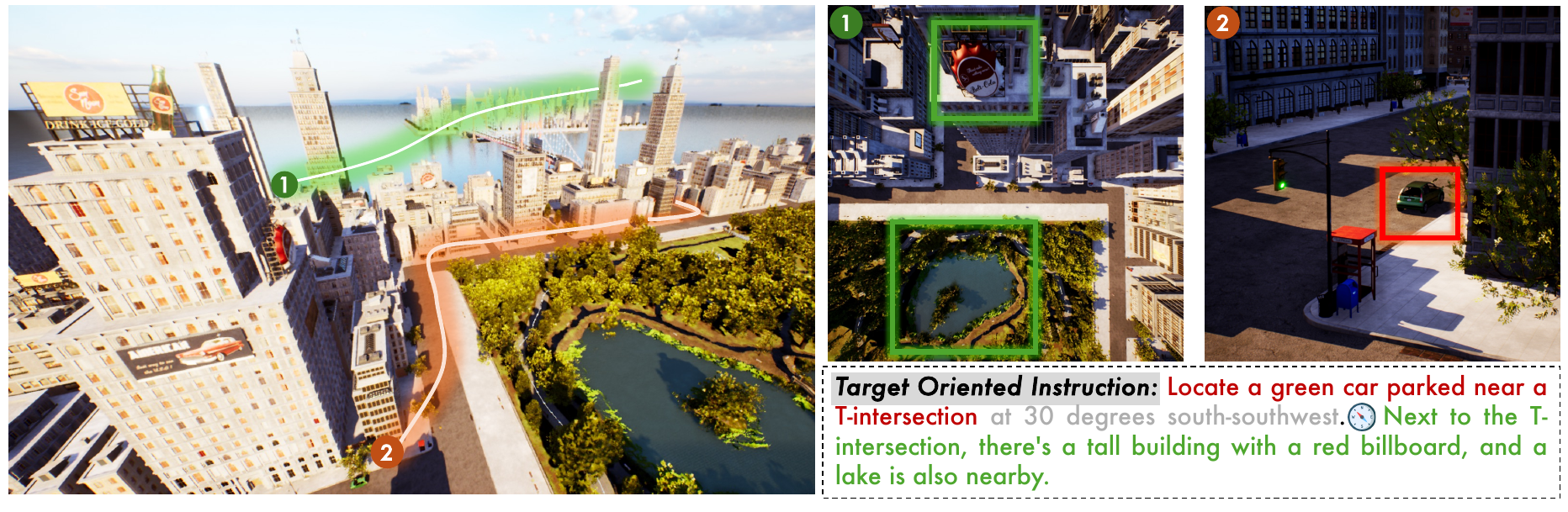}
    \vspace{-3mm}
    \captionof{figure}{
    In the DuAl-VLN task, the two UAVs operating at distinct altitudes achieve collaborative target search through language instruction guidance. The high-altitude UAV offers a broader observation range (\protect\scalerel*{\includegraphics{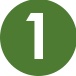}}{B}), while the low-altitude UAV captures finer-grained visual perception of target (\protect\scalerel*{\includegraphics{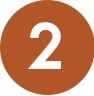}}{B}).
}
\label{fig:fig1}
\end{teaserfigure}

% 删除 teaser



\maketitle

\section{Introduction} \label{sec:intro}

Vision-Language Navigation (VLN)~\cite{Fried_Hu_Cirik_Rohrbach_Andreas_Morency_Berg-Kirkpatrick_Saenko_Klein_Darrell_2018}, which aims to enable autonomous agents to navigate based on natural language instructions, has recently received significant research attention. Early efforts primarily focused on ground-based agents and have achieved remarkable progress. In contrast, VLN for Unmanned Aerial Vehicles (UAVs) remains relatively understudied. UAV-based VLN poses greater challenges due to extended navigation trajectories and higher degrees of motion freedom compared to ground-based scenarios. Existing approaches~\cite{liu2023aerialvln,wang2024realisticuavvisionlanguagenavigation,lee2024citynav} for UAV-based VLN often rely on iterative human-agent dialogues~\cite{fan2022aerial}, highly detailed route descriptions~\cite{gao2024aerial}, or real-time human assistance~\cite{wang2024realisticuavvisionlanguagenavigation}, which inevitably increase human workload and restrict operational efficiency. Consequently, enabling UAVs to accomplish VLN tasks solely through relatively simple instructions, which only describe target orientation, regions, and surrounding features, could significantly enhance practicality. However, this remains challenging for a single UAV agent, as it struggles to simultaneously achieve high-altitude perspective for coarse-grained regional localization and low-altitude perspective for fine-grained target observation.

To tackle this challenge, we introduce a novel UAV-based VLN task called Dual-Altitude UAV Collaborative Vision-Language Navigation (DuAl-VLN). 
In this task, two UAVs operate at different altitudes: one at high altitude for wide-area environmental coverage and the other at low altitude for detailed, close-range observation. The two UAVs collaboratively optimize navigation through dynamic information exchange and joint path planning, leveraging their complementary capabilities: high-altitude contextual awareness enhances strategic decision-making, while low-altitude detailed sensing ensures real-time obstacle avoidance and safe navigation in cluttered environments. This dual-altitude framework significantly enhances navigation efficiency by balancing macro-scale environmental understanding with micro-scale flight safety.

To advance this task, we curate a dataset, HaL-13k, based on the OpenUAV platform~\cite{wang2024realisticuavvisionlanguagenavigation} with $13,838$ synchronized dual-altitude trajectories across 14 scenarios. This dataset is generated using a two-stage trajectory creation process: first, low-altitude UAV trajectories are obtained via obstacle-aware path planning; subsequently, high-altitude UAV paths are collected under strict visibility constraints. The constraints ensure visual overlap with the low-altitude trajectories, thereby enabling effective autonomous exploration for high-altitude UAVs.
HaL-13k offers navigation instructions that exclusively describes target orientation, visual features, and surrounding environmental context, along with paired high-low trajectories and multi-modal sensor streams.
This dataset is tailored to investigate altitude-dependent perception dynamics and collaborative decision-making in the dual-UAV system.

To address the challenges of cooperative navigation, where inefficient information exchange can cause trajectory conflicts, we propose AeroDuo. This collaborative UAV-VLN framework synergizes multimodal large language models and lightweight models tailored for each UAV's role. \textbf{For the high-altitude UAV}, we introduce the Pilot-LLM, which leverages pre-trained MLLMs' capabilities to enable effective instruction understanding and target reasoning. Specifically, Pilot-LLM processes historical flight trajectories and constructs a global orthographic projection map to dynamically infer coarse-grained target regions. A mask prediction module is further integrated into the MLLM, prioritizing feasible areas for the low-altitude UAV's detailed exploration.
\textbf{For the low-altitude UAV}, we deploy a navigation policy trained in the Isaac Sim~\cite{makoviychuk2021isaac} simulation environment, combining a lightweight obstacle avoidance controller and a visual grounding model to precisely localize target objects. Crucially, the two UAVs communicate only minimal coordinate information, significantly reducing bandwidth requirements while maintaining collaborative coherence.
Experimental results demonstrate that our AeroDuo achieves a significant improvement of \textbf{9.71\%} in navigation success rates compared to single-UAV baselines on the validation set of the HaL-13k dataset, which demonstrates the effectiveness of dual-altitude UAV collaboration for VLN tasks, opening new possibilities for aerial embodied AI systems.
\section{Related Works}
\subsection{Ground-based Vision-Language Navigation}
Ground-based VLN has seen rapid advances, with datasets~\cite{anderson2018vision,Matterport3D} and benchmarks~\cite{qi2020reverie,jain2019stay,ku2020room,chen2019touchdown,krantz2020beyond} enabling broader task coverage through diverse instructions and heterogeneous environments.
Related research has delved into data augmentation techniques~\cite{Fried_Hu_Cirik_Rohrbach_Andreas_Morency_Berg-Kirkpatrick_Saenko_Klein_Darrell_2018, Zeng_Wang_Wang_Yang_2023, wang2022less, Zhang_Kordjamshidi_2023}, decision-making mechanisms~\cite{zhao2022target}, the utilization of historical context~\cite{chen2021history, hong2020recurrent,gao2021room,kong2024controllable,chen2022reinforced,gao2023room}, and representations of three-dimensional space. Furthermore,
the rapid advancement of LLMs~\cite{vicuna,llama} and MLLMs~\cite{mlvu,guo2025llava,li2025llavastmultimodallargelanguage,timechat} has inspired action-prediction methods like ~\cite{brohan2023rt}. Recent works~\cite{li2024panogen, chen2024mapgpt,qiao2024llm} integrate LLMs into planning, while others~\cite{zheng2024towards,zhang2024navid} propose unified models for language and environmental context. Compared to ground-based VLN, UAV-based VLN exhibits a longer trajectory length and higher degrees of motion freedom, presenting significantly greater challenges.

\subsection{UAV Navigation}
Current UAV navigation research has mainly focused on visual perception~\cite{loquercio2018dronet,giusti2015machine,smolyanskiy2017toward,fan2020learn,Ilker2020multimodal,Amdras2017Zurich,kang2019generalization} and collision avoidance~\cite{xu2025navrl,singla2019memory}, while multimodal visual-linguistic UAV navigation remains emerging.
Recent efforts~\cite{liu2023aerialvln,liu2024navagent} introduced UAV-VLN frameworks using detailed textual guidance. AerialVLN~\cite{liu2023aerialvln} provides a large-scale dataset with an effective baseline, and STMR~\cite{gao2024aerial} builds on it with a zero-shot LLM-based framework using a Semantic-TopoMetric Representation for spatial reasoning.
Other works have contributed infrastructure datasets, including CityNav~\cite{lee2024citynav}, AVDN~\cite{fan2022aerial}, and OpenFly~\cite{gao2025openfly}. OpenUAV~\cite{wang2024realisticuavvisionlanguagenavigation}, in particular, offers a UAV dynamics simulator and the UAV-Need-Help evaluation protocol. Despite these advances, current UAV-based VLN research has not explored multi-agent collaboration for enhanced navigation performance.

\subsection{Multi Agent Navigation}
Multi-agent collaboration has been widely explored, with early work focusing on reinforcement learning for  autonomous coordination in structured environments~\cite{perez2019multi,lowe2017multi,samvelyan2019starcraft,berner2019dota}. Recent advances leverage LLMs to assign roles via prompts, enabling language-based interaction~\cite{hongmetagpt}.
In ground-based navigation, studies explore multi-agent VLN in indoor environments~\cite{zhu2024communicative} and cooperative target search in games~\cite{zhao2024hierarchical}.
By contrast, UAV-focused research centers on swarm formation and obstacle avoidance~\cite{pant2018fly,2021done1multi}, without addressing multimodal understanding. However, Multi-agent VLN for UAVs remains unexplored, further challenged by UAVs’ large operational space and degrees of motion freedom.

\section{Dual-Altitude UAV Collaborative Vision--Language Navigation}
In this paper, we introduce a novel UAV-based VLN task, Dual-Altitude UAV collaborative VLN (DuAl-VLN), that leverages dual-altitude UAVs to achieve collaborative perception and decision-making. This task strategically balances UAVs' inherent high mobility with constrained operational spaces, creating an optimized environment for model learning while preserving aerial maneuverability advantages. We detail the task setup of the DuAl-VLN in Sec.~\ref{sec:setup}. To support this task, we develop the first dual-UAV VLN dataset, HaL-13k, which provides concise instructions of targets, featuring dual-altitude trajectories with multi-modal sensor data (Sec.~\ref{sec:dataset}).
\begin{figure}
    \centering
    \includegraphics[width=0.95\linewidth]{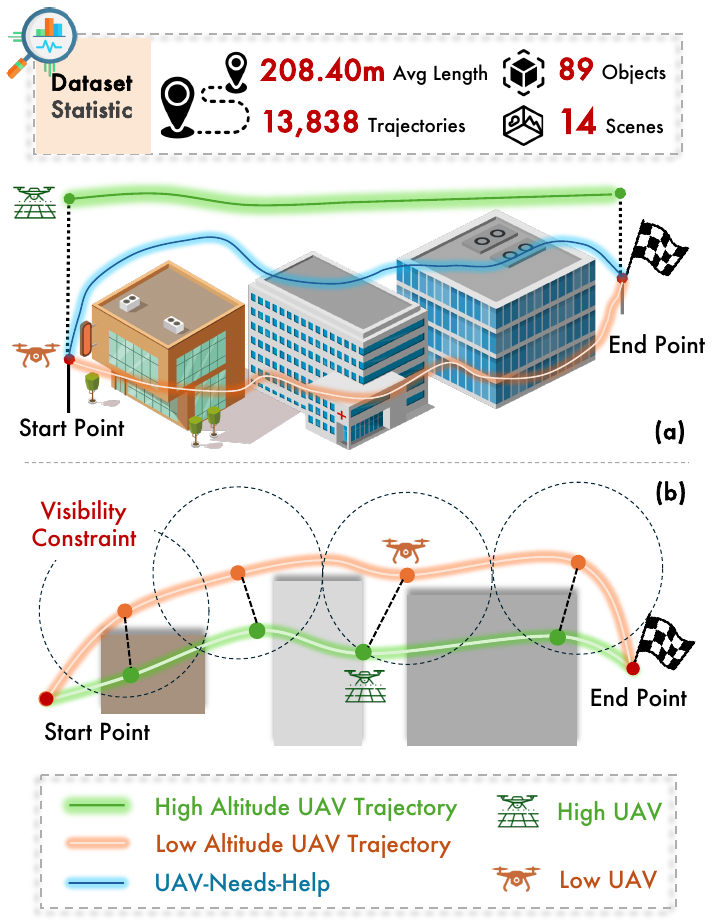}
    \vspace{-2mm}
    \caption{Illustration of dataset statistics and trajectory collection process for HaL-13k. (a) We optimize the paths from UAV-needs-Help~\cite{wang2024realisticuavvisionlanguagenavigation} to maintain an effective exploration altitude for the low-altitude UAV. (b) We randomly sample paths of the high-altitude UAV that obey visibility constraints, ensuring that the high-altitude UAV maintains visual coverage of the low-altitude UAV’s route.}
    \label{fig:dataset}
    \vspace{-2mm}
\end{figure}

\subsection{Task Formulation}
\label{sec:setup}
At the beginning of each episode, the low-altitude UAV $ U_l $ and the high-altitude UAV $ U_h $ are initialized at positions $ P^l_0 = (x^l_0, y^l_0, z^l_0) $ and $ P^h_0 = (x^h_0, y^h_0, z^h_0) $, respectively, with $ x^l_0 = x^h_0 $, $ y^l_0 = y^h_0 $, and $ z^h_0 > z^l_0 $. The dual-UAV system receives a target-oriented linguistic instruction describing the target's direction, characteristics, and surrounding environmental context. To reflect the difference in reasoning frequency, we denote the decision time steps for the low-altitude UAV $U_l$ as $t$ and for the high-altitude UAV $U_h$ as $\tau$. Specifically, at each time step $t$, $U_l$ captures forward-facing visual image $I^l_t$ and omnidirectional point cloud data $V^l_t$. At each time step $\tau$, $U_h$ captures a BEV observation $I^h_\tau$ along with a Lidar point cloud map $V^h_\tau$, both covering the same field of view. Leveraging these multi-modal inputs, the UAVs dynamically adjust their flight trajectories by predicting either subsequent waypoint sequences or velocity profiles supported by Airsim~\cite{shah2018airsim}. A navigation episode is deemed successful if $U_l$ comes within a distance threshold $d$ of the target location $p^d$. A navigation episode is considered failed if either the $U_l$ exceeds the navigation-time upper-bound without reaching the target or the UAVs collide with obstacles.
\begin{figure*}
    \centering
    \includegraphics[width=0.96\linewidth]{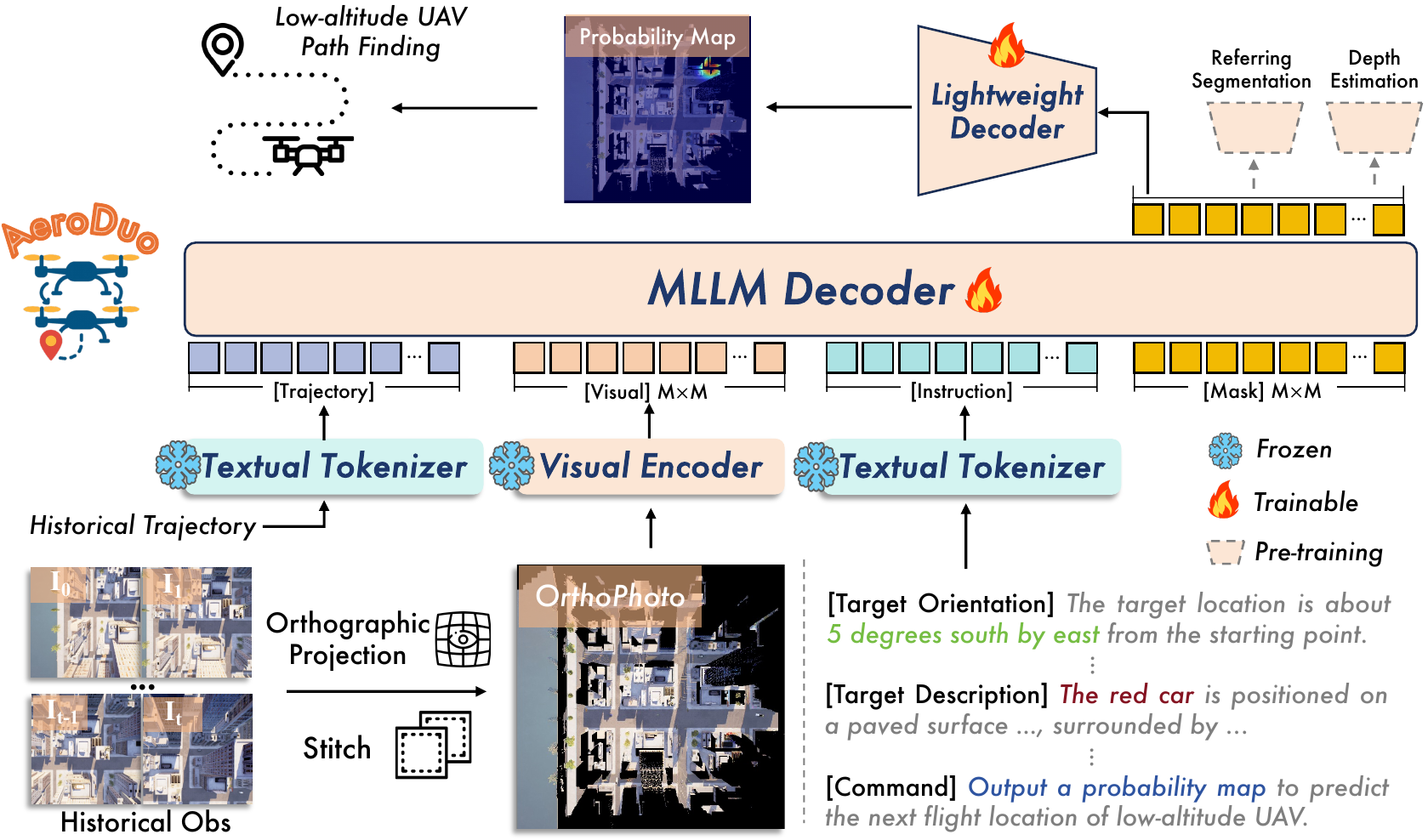}
    \caption{The AeroDuo framework integrates multiple data sources to achieve precise UAV navigation. We input a global orthophoto map, UAV historical trajectories, and linguistic instructions into the MLLM decoder, \highllm. Special mask tokens are employed to predict the probability distribution of target locations.
    The flight target is obtained from the resulting probability map, which is subsequently used for pathfinding by low-altitude UAV. To enhance the geospatial modeling capabilities of the MLLM, we pretrain it using auxiliary tasks such as referring segmentation and depth estimation.}
    \label{fig:method}
\end{figure*}

\subsection{HaL-13k Dataset}
\label{sec:dataset}
To advance the DuAl-VLN task, we deliberately construct a dataset, HaL-13k, upon the OpenUAV~\cite{wang2024realisticuavvisionlanguagenavigation} platform, which could provide realistic UAV sensory data, diverse environments, and dynamic flight characteristics. However, existing RL-based multi-agent collaboration approaches, which rely on frequent environmental interactions, impose prohibitive computational costs for simulating. To overcome this limitation, we propose to collect expert continuous trajectory data for coordinated high-low UAV pairs. To build the trajectory pairs, we optimize the navigation paths from \cite{wang2024realisticuavvisionlanguagenavigation} to maintain an effective exploration altitude for the low-altitude UAV, while generating trajectories with optimal perspective for the high-altitude UAV.
Specifically, for the low-altitude UAV path planning, we first construct an occupancy map from point clouds. The A* algorithm~\cite{hart1968formal} is then used to compute optimized navigation paths that conform to these occupancy constraints. For the high-altitude UAV, we randomly sample flight paths that guarantee full visual coverage of the low-altitude UAV's route, thereby maintaining optimal observational perspectives throughout the mission.

As shown in Fig.~\ref{fig:dataset}, this generation strategy yields a dataset of $13,838$ collaborative trajectory pairs, each annotated with target-oriented language instructions for search target, without highly detailed route descriptions~\cite{gao2024aerial}, or real-time human assistance~\cite{wang2024realisticuavvisionlanguagenavigation}.

For the validation set of HaL-13k,  we partition it into the unseen map set and the unseen object set, ensuring rigorous evaluation of generalization to novel layouts and unfamiliar objects.
\begin{itemize}[leftmargin=1em]
    \item \textbf{Unseen Map Set} consists of environments absent from the training data. We sample 2 scenes as unseen maps and extract all associated trajectories for testing, resulting in 175 episodes. 
    \item \textbf{Unseen Object Set} comprises trajectories from training-familiar scenes but introduces object categories never encountered during training, yielding 175 test episodes.  
\end{itemize}

\section{AeroDuo: Aerial Duo for UAV-based VLN}\label{sec:method}

\subsection{Overview of AeroDuo}
We propose a dual-altitude UAV collaborative VLN framework, \textbf{AeroDuo}, which integrates MLLMs for high-altitude decision making with lightweight multi-stage policies for low-altitude navigation. In this section, we will provide a detailed overview of each component of AeroDuo. At timestep \(\tau\), following an exploration request from the low-altitude UAV \(U_l\), the high-altitude UAV \(U_h\) initiates its decision-making phase. For clarity, we denote \(U_h\)'s decision timesteps as \(\tau\) in the following part. During this phase, \(U_h\) employs an MLLM-based decoder, \highllm, to predict a target probability map \(\mathcal{M}_\tau\) and generates corresponding environmental depth information \(\hat{D}^h_\tau\). These outputs are then transmitted to \(U_l\). Equipped with \(\mathcal{M}_\tau\) and \(\hat{D}^h_\tau\), \(U_l\) performs environmental exploration using the Multi-Stage Pathfinder (\lowuav). Upon completing the exploration, \(U_l\) requests the next target from \(U_h\) to continue the navigation process. Detailed descriptions of \highllm and \lowuav are provided in Sec.~\ref{sec:highUAV} and Sec.~\ref{sec:lowUAV}.

\subsection{\highllm on High UAV}
\label{sec:highUAV}

The primary advantage of high-altitude UAVs lies in their wide-field observation capability, which enhances the system’s efficiency in both pinpointing target areas and planning subsequent navigation paths.
Previous methods process UAV observations as video frames~\cite{liu2023aerialvln,wang2024realisticuavvisionlanguagenavigation}, which usually suffer from two critical limitations. First, video sequences prioritize temporal ordering of exploration snapshots, which obscures the spatial relationships between historical observations. Consequently, when key regions span multiple frames, the model would struggle to determine the target location. Second, the lack of explicit spatial coordinate cues in video sequences hinders precise environmental mapping, resulting in unreliable location predictions and compromised results. To alleviate these issues, we propose constructing a global map that integrates historical observations, thereby offering a holistic perspective and a unified spatial coordinate system for the dual-UAV system.

% Previous methods processed UAV observations as video frames~\cite{liu2023aerialvln,zheng2024towards}, but this approach has two significant drawbacks. First, video sequences make it difficult for the model to accurately understand the spatial relationships between different historical observations, as they only describe temporally ordered observations during exploration. Consequently, when a critical area is split across two video frames, the model struggles to determine the target location. Second, this representation lacks effective spatial coordinate information, leading to inaccurate predictions of environmental coordinates.  
\paragraph{Global Map Construction.}
Although the high-altitude UAV captures RGB observations in a BEV-like fashion, directly stitching these views into a coherent global map remains non-trivial due to perspective distortions.
To address this issue, we employ an orthophoto generation pipeline. Specifically, we first reconstruct an elevation map of the ground environment from the accumulated point clouds $V^h_{1:\tau}$. Then, using the UAV's trajectory $p^h_{\tau}$, we estimate the extrinsic parameters of the onboard RGB camera and reproject the BEV observations $I^h_{1:\tau}$ to the ground plane. These reprojected views are then stitched together to produce the global orthographic map $G_\tau$.
Meanwhile, the current UAV position $p^h_\tau$ is also used to compute the global depth map $\hat{D}^h_\tau$ with respect to the UAV's camera plane. The overall process is summarized as:
\begin{equation}
    G_{\tau}, \hat{D}^h_\tau= ortho(I^h_{1:\tau},p^h_{1:\tau},V^h_{1:\tau}).
\end{equation}
We limit the stitching to a maximum of five historical images. The orthophoto map will be resized to a pre-defined size and then fed into the \highllm as an integrated historical observation. Please refer to the supplementary material for more details.

\paragraph{\highllm.} 
To enable effective perception and decision making with multimodal inputs for the dual-UAV system, we take advantage of the reasoning ability of multimodal large language models (MLLMs) to handle various input types and generate pilot guidance in a unified framework, \highllm.

At time step $\tau$, the input to \highllm consists of three key components: the orthophoto map $G_{\tau}$, the historical trajectory $p^h_{1:\tau}$, and the navigation instruction.
For the orthophoto map $G_{\tau}$, we tokenize it into visual tokens $\tilde{\bm{G}}{\tau} \in \mathbb{R}^{N \times D}$ using the visual encoder $f_v$, where $N$ and $D$ represents the number and the dimension of visual tokens. For the historical trajectory $p^h_{1:\tau}$, we first project it onto the coordinate system of $G_{\tau}$, denoted as $\hat{p}^h_{1:\tau}$, and then encode these positions into trajectory embeddings $\bm{X}^p_\tau$ using the textual tokenizer.
The navigation instruction $X$ is also processed through the LLM's textual tokenizer to generate embeddings $\tilde{\bm{X}}$.
These tokens are then flattened, concatenated, and subsequently fed into the \highllm.

Given these inputs, \highllm ought to predict the precise target location for the dual-UAV system.
However, directly predicting spatial coordinates of target locations in textual format would result in significant errors, because LLMs struggle with explicit geospatial modeling, as highlighted in~\cite{yamada2023evaluating,bhandari2023large}. Instead of outputting precise coordinates, we propose to predict a probability distribution map $G_\tau$ that emphasizes candidate target regions. This design offers two clear advantages. First, the flight target for UAVs should represent a feasible area rather than a single coordinate point, thereby preserving their exploration ability. Second, by predicting a map, the spatial modeling capacity of \highllm can be enhanced by incorporating auxiliary tasks, such as referring segmentation and depth estimation, as detailed later in Sec.~\ref{sec:LLM-training}.

Formally, to generate a distribution map by \highllm, we incorporate learnable special tokens $\bm{M} \in \mathbb{R}^{N \times D}$, where each token corresponds to a unique spatial coordinate in the orthophoto map $G_{\tau}$.
The embedding at the coordinate $(i,j)$ in $\bm{M}$ is defined as:  
\begin{equation}
 \bm{M}(i,j) = \bm{\rho}_{i,j} + \bm{\eta},  
\end{equation}
where $ \bm{\rho}_{i,j}\in R^{1\times D}$ denotes the positional embedding at coordinate $(i,j)$ in $f_v$ and $\bm{\eta}$ is a trainable embedding. Here, we use the same coordinate representation as $ f_v $ to ensure accurate mask prediction.  
Finally, \highllm takes $[ \tilde{\bm{G}}_{\tau},\bm{X}^p_{\tau},\tilde{\bm{X}},\bm{M}]$ as input, and the output features $\tilde{\bm{M}}_{\tau}$ corresponding to the $ \bm{M} $ are decoded by a lightweight mask decoder $f_m$ to predict the probability of target location:
\begin{equation}
    \mathcal{M}_{\tau}=\text{sigmoid}(f_m(\tilde{\bm{M}}_{\tau})),
\end{equation}
where $\mathcal{M}_\tau$ is a target location probability map with the same spatial size as $G_{\tau}$. After that, the low-altitude $U_l$ will take the $\mathcal{M}_\tau$ and the global depth map $\hat{D}^h_{\tau}$ for explorative path finding. 

For $U_h$, its high flight altitude mitigates collision risks while offering an extensive observational field, enabling efficient surveillance of target-proximate areas through orientation-optimized directional movement. During the navigation, $U_h$ first computes the flight direction and step length using instructions and compass data, then follows the predictions throughout subsequent operations.

\subsection{Multi-Stage Pathfinder on Low UAV}
\label{sec:lowUAV}
\begin{figure}
    \centering \includegraphics[width=1\linewidth]{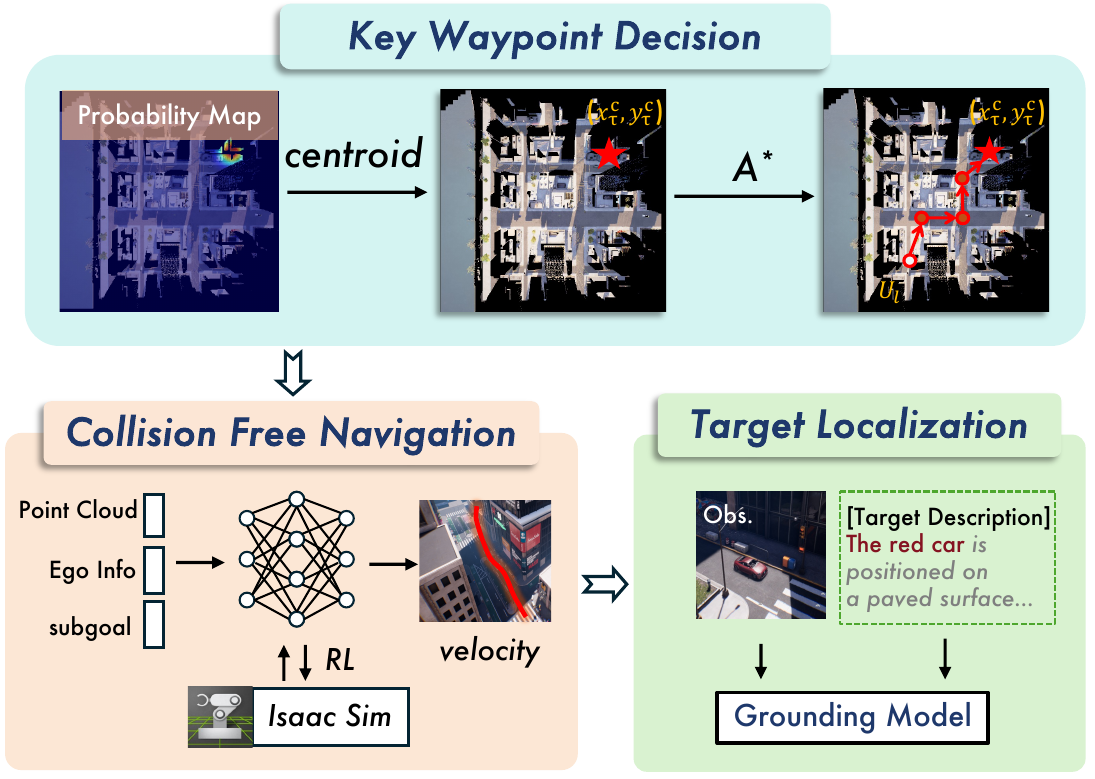}
    \caption{Overview of Multi-Stage Pathfinder (\lowuav) on low-altitude UAV, including successive stages of Key Waypoint Decision, Collision-Free Navigation and Target Localization.}
    \label{fig:lowuav}
    \vspace{-3mm}
\end{figure}

Upon completion of environmental mapping and target probability estimation by the high-altitude UAV $U_h$, the low-altitude UAV $U_l$ performs explorative navigation guided by $U_h$ to find the target instance. To enable a safe and efficient environment exploration, we propose a Multi-Stage Pathfinder (\lowuav) for $U_l$. As shown in Fig.~\ref{fig:lowuav}, the \lowuav pipeline executes navigation through three core stages: key waypoint decision, collision-free navigation, and target localization.

\paragraph{Key Waypoint Decision.} 
In this stage, the low-altitude UAV, located at \(p^l_t = [x^l_t, y^l_t, z^l_t]\), first determines its sub-goal by computing the centroid of the probability distribution \(\mathcal{M}_\tau\) to effectively mitigate errors caused by outliers:
\begin{equation}
    [x^c_{\tau},y^c_{\tau}] = \sum_{i}^H\sum_{j}^H \mathcal{M}_{\tau}(i,j) \cdot [i,j].
\end{equation}
Since the high-altitude and low-altitude UAVs start from the same horizontal coordinate (albeit at different altitudes) and the trajectory of the high-altitude UAV is known, it is straightforward to transform the coordinates 
$[x^c_{\tau}, y^c_{\tau}]$
into the global coordinate system. This transformation results in the predicted endpoint 
$[\hat{x}^c_{\tau},\hat{y}^c_{\tau}, z_t^l]$. With the sub-goal established, the UAV generates a sequence of key waypoints \(\mathcal{Q}_\tau\) to navigate toward the target. During this process, leveraging \(U_h\)’s wide-range perspective, which provides a comprehensive understanding of the environmental context, would significantly improve \(U_l\)’s exploration efficiency. To achieve this, we first construct the occupancy map \(occ_\tau\) based on the global depth map \(\hat{D}^h_\tau\). An approximate navigation path is then derived by optimizing \(occ_\tau\) using the A* algorithm~\cite{hart1968formal}. The occupancy map \(occ_\tau\) is calculated as follows: 
\begin{equation}
    occ_\tau = u(\hat{D}^h_\tau - \Delta z_\tau),
\end{equation}
where $\Delta z_\tau$ denotes the altitude difference between the high- and low-altitude UAVs, and $u(\cdot)$ denotes the unit step function as:
\begin{equation}
u(x) = \mathbf{1}_{{x \geq 0}}.
\end{equation}
The A* search algorithm is performed on \( occ_\tau \) using the Manhattan distance as the heuristic function to search a trajectory from $p^l_t$ to $p^c_\tau$. Furthermore, an erosion operation is applied to the occupancy map to allow the A* algorithm to search for paths at a safer distance from obstacles. Based on the length of the trajectory, the trajectory is equally segmented into \(K\) key waypoints, denoted as \( \mathcal{Q}_\tau = \{ p^\tau_{1:K} \} \).

\paragraph{Collision-Free Navigation.}
Due to the low spatial resolution of $occ_\tau$, relying solely on the computed waypoints \( \mathcal{Q}_\tau = \{ p^\tau_{1:K_\tau} \} \) from $\hat{D}_\tau^h$ often leads to collisions. To mitigate this issue, we employ an RL-based collision-free Navigator, inspired by~\cite{xu2025navrl}. At timestep \( t \) of $U_l$, the Navigator receives the point cloud \(\mathbf{V}^l_t\), the subgoal $p^\tau_k \in \mathcal{Q}_\tau$, and the ego status, which includes the current position $p^l_t$ and the current velocity $v^l_t$, for the 
next timestep's velocity prediction. To efficiently encode \(\mathbf{V}^l_t\), we adopt a 3D ray-casting strategy. 
Thereafter, we feed the encoded point clouds $\hat{V}^l_{t}$ and the other inputs into a multi-layer perceptron (MLP) to predict the subsequent velocity $v^l_{t+1}$. The network employs the PPO~\cite{schulman2017proximal} algorithm for training, with a reward function that incorporates obstacle avoidance, penalties for velocity fluctuations, and incentives for reducing the distance to the target as in ~\cite{xu2025navrl}.  
To accelerate the training process, we employed Isaac Sim~\cite{makoviychuk2021isaac} as the training simulator, leveraging its high-speed parallel simulation. Since the navigator relies solely on point cloud data, it facilitates seamless adaptation across different simulated environments and real-world scenarios. More details are presented in the supplementary material.

\paragraph{Target Localization.} 
During navigation along \(\mathcal{Q}_\tau\), the low-altitude UAV \(U_l\) continuously searches for the target and terminates its exploration once the target is successfully detected. Otherwise, if \(U_l\) reaches the end of \(\mathcal{Q}_\tau\) without detection, it will request new navigation guidance from \(U_h\). 
We follow~\cite{wang2024realisticuavvisionlanguagenavigation} to adopt GroundingDINO~\cite{liu2023grounding} as the detector \(f_g\), enabling target localization based on textual instructions. The visual grounding process operates asynchronously with navigation: after each detection attempt is completed, \(f_g\) is immediately applied to the latest observation from \(U_l\). The exploration terminates once the confidence score of the detected bounding box exceeds a predefined threshold.

\subsection{Training Process of \highllm}
\label{sec:LLM-training}

\begin{table*}[!t]
\centering
\caption{The main comparison on the validation set of HaL-13k. For fair comparison, we train these baseline methods on the proposed dataset of HaL-13k.}
% \vspace{-2mm}
\setlength{\tabcolsep}{3.0mm}{
\begin{tabular}{lcccccccccc}
\toprule
         \multirow{2}{*}{\textbf{Method}}   & \multicolumn{5}{c}{\textbf{Unseen Map}} & \multicolumn{5}{c}{\textbf{Unseen Object}}  \\
         \cmidrule(r){2-6} 
         \cmidrule(r){7-11}
& SST $\uparrow$ & SR $\uparrow$ & SPL $\uparrow$ & OSR $\uparrow$ & NE $\downarrow$ & SST $\uparrow$ & SR $\uparrow$ & SPL $\uparrow$ & OSR $\uparrow$ & NE $\downarrow$ \\
\midrule
Random     &  0.00  & 0.00   &  0.00   &  0.00   &   199.42     &    0.00 &  0.00  &  0.00   &  0.00   &  199.25          \\
CMA~\cite{liu2023aerialvln}        &  0.00   & 0.00   & 0.00   &  0.57   &   166.31      &  0.00   &  0.00  &  0.00   &    0.57 &  179.30           \\
TravelUAV~\cite{wang2024realisticuavvisionlanguagenavigation} &  0.57   & 0.57  &  0.57 &   1.14  &    152.20     &   0.56  &  0.57  &  0.52   &   2.86  &    160.16            \\
TravelUAV (L1 assistant)     &   6.48  &  6.86  &   5.89 & 17.14 &  107.91  &  5.31   & 5.71   &  5.05   &  10.29   &   140.42         \\
\rowcolor{gray!10} AeroDuo    & \textbf{14.63}  &   \textbf{16.57}  &  \textbf{13.86}  &  \textbf{28.57}   &  \textbf{84.31}   &    \textbf{13.54}  &    \textbf{14.86} &  \textbf{13.35}  &   \textbf{19.43} &   \textbf{108.66} \\
\bottomrule         
\end{tabular}}
\label{tab:main_comparison}
\end{table*}

In this section, we present the training pipeline of \highllm, which consists of a pretraining stage and a finetuning stage. The pretraining stage aims to enhance the general visual and geospatial modeling ability of \highllm. In the fine-tuning stage, \highllm is trained to localize the exploration areas based on BEV observations.

\paragraph{Pretraining Stage.}
Since \highllm needs to locate target regions based on navigation instructions, it requires strong cross-modal understanding capabilities. While general MLLMs provide a solid foundation for this task, they lack two critical capabilities:  
First, the observations $ U_h $ consist of BEV images. Due to the limited amount of BEV data in the general MLLM training, existing models struggle to cross-modal understanding from a BEV perspective. Second, predicting key navigation regions requires robust spatial reasoning ability, which helps to understand whether structures or environmental elements may hinder navigation. To enhance these capabilities, we introduce a pretraining phase for \highllm. (a) \textit{To boost the model's general perception ability}, we first train the Pilot-LLM on referring segmentation and depth estimation based on the RefCOCO dataset~\cite{yu2016modeling}. In our method, we employ the Depth Anything v2~\cite{yang2024depth} to generate depth maps for training. 

After training on the general dataset, (b) \textit{to enhance the ability of BEV image understanding}, we train the model using referring segmentation on BEV images. Multiple text templates are then used to generate descriptive captions for these objects, resulting in a dataset of $850,894$ image-text pairs. During training, the predicted mask is generated by $ f_m $ using $\tilde{\bm{M}}$. 
(c) \textit{To improve the model’s geospatial perception}, we train the Pilot-LLM for depth estimation from BEV images. A separate decoder is employed to ease learning difficulty. 

Notably, we train the Pilot-LLM on mixed data from both tasks.
This pretraining strategy ensures \highllm develops both BEV-aware vision-language alignment and precise spatial reasoning for navigation. More details are shown in the supplementary materials.

\paragraph{Finetuning Stage.} After the pretraining stage, the \highllm needs to further locate the key exploration area based on language instructions. 
% Since VLN navigation data lacks annotations~\cite{wang2024realisticuavvisionlanguagenavigation} for key regions and only includes navigation target point coordinates, using a single coordinate point to supervise mask generation would significantly increase training difficulty. 
Here, we first initialize the mask decoder with the parameters of the segmentation decoder in the pre-training stage, as referring segmentation and mask prediction are similar tasks.
To generate the ground truth labels for training, we sample a pair of high-low UAV waypoints at a given time step \( t \). The low-altitude UAV's position at a future time step \( t + k \) is used as the target. A Gaussian distribution is then centered at this future position to produce a probability distribution map. If the target position at \( t + k \) falls outside the orthophoto map, we instead assign the nearest valid point within the map as the surrogate target position. After that, an occupancy mask is applied to suppress infeasible areas by setting their probabilities to zero. Finally, the probability map is normalized to obtain the final ground truth label for the low-altitude UAV. We employ a similar strategy to obtain the ground truth label for the high-altitude UAV, with the objective shifted to predicting its heading and step length.

\section{Experiments}
\begin{table*}[!ht]
\centering
\caption{The ablation study on the validation set of HaL-13k.}
\vspace{-2mm}
\setlength{\tabcolsep}{1.3mm}{
\begin{tabular}{>{\columncolor{gray!10}}c>{\columncolor{gray!10}}c>{\columncolor{gray!10}}c>{\columncolor{gray!10}}ccccccccccccc}
\toprule
\multicolumn{4}{>{\columncolor{gray!10}}c}{\textbf{Method}} & \multicolumn{2}{c}{\textbf{Average}} & \multicolumn{5}{c}{\textbf{Unseen Map}} & \multicolumn{5}{c}{\textbf{Unseen Object}}   \\
\cmidrule(r){1-4}
\cmidrule(r){5-6}
\cmidrule(r){7-11}
\cmidrule(r){12-16}
     Pretrain  & GMC  &KWD &CFN & SST $\uparrow$ & SR $\uparrow$ & SST $\uparrow$ & SR $\uparrow$ & SPL $\uparrow$ & OSR $\uparrow$ & NE $\downarrow$ & SST $\uparrow$ & SR $\uparrow$ & SPL $\uparrow$ & OSR $\uparrow$ & NE $\downarrow$                              \\
\midrule

    &  & & $\checkmark$ &  1.34 &  1.43  &  1.71   &   1.71  &    1.62 &  3.43   &  146.08  &  0.97   &  1.14   &    0.98     &   1.71  &   199.78                     \\
$\checkmark$  &  &  & $\checkmark$ & 2.08 &  2.29  &  3.15  &  3.43  &  2.97       &  6.29   &  107.95  &  1.02   &     1.14  & 0.99 &  4.57  &   149.99                     \\
$\checkmark$  & $\checkmark$ &  & $\checkmark$&  12.75   & 14.29 &   \textbf{17.27}   & \textbf{19.43}   &   \textbf{16.44}  &   \textbf{32.00}  &    \textbf{78.55}   &  8.24   &  9.14  &  8.12   & 10.86 & 126.80        \\
$\checkmark$  & $\checkmark$ & $\checkmark$  &  &    8.08  & 8.86 &      8.57  &  9.14   &  8.13 &   24.57      &    98.89 &  7.59  &  8.57   &  7.30   &     11.43       &  135.40    \\             
$\checkmark$  & $\checkmark$ & $\checkmark$ &  $\checkmark$ &  \textbf{14.08}   &  \textbf{15.71}  &  14.63  &   16.57  &  13.86  &  28.57   &  84.31   &    \textbf{13.54}     &    \textbf{14.86} &  \textbf{13.35}  &   \textbf{19.43} &  \textbf{108.66}    \\               
\bottomrule         

\end{tabular}}
\label{tab:ablation_study}
\end{table*}
\subsection{Experiment Setup}
\paragraph{Implementation Details.} 
Our \highllm framework builds upon the visual projector \( f_v \) and LLM backbone from Qwen2-VL~\cite{Qwen2-VL}. The mask prediction head consists of two linear layers followed by two upsampling layers, aiming to expand the output resolution.  
For optimization, we employ the AdamW~\cite{loshchilov2017decoupled} optimizer with a cosine learning rate scheduler, initialized at \( 5 \times 10^{-5} \). During training, we freeze the visual projector and fine-tune the MLLM using the low-rank adaptation (LoRA). The mask and depth prediction tasks are optimized using binary cross-entropy loss and MSE loss, respectively.  
The low-altitude UAV \( U_l \) operates at a control frequency of 10 Hz, with each time step \( t \) spanning 0.1 seconds.

\vspace{1mm}
\paragraph{Evaluation Metrics.}
We utilize the following 5 metrics to evaluate the performance of the navigation model:
\begin{itemize}[leftmargin=1em]
    \item  \textbf{SR}: Success Rate. SR measures the percentage of tasks in which the UAV successfully halts within a \(20\text{ m}\) radius of the target.
    \item \textbf{SPL}: Success rate weighted by Path Lengh. SPL combines task success with path efficiency by multiplying success rate by the ratio of optimal to actual path length of the low-altitude UAV.
    \item \textbf{SST}: Success rate weighted by Search Time. SST measures navigation time efficiency. It is calculated as:
    % \begin{equation}
    %     SST = S \times ({T^*} \div {\max(T, T^*)})
    % \end{equation}
    $SST = S \times ({T^*} \div {\max(T, T^*)})$
    where $T$ is the target search simulator time, and $T^*$ is the navigation time for the ground-truth trajectory.
    % \item \textbf{CR}: Collision Rate. CR measures the proportion of episodes where the UAV aborts navigation due to collisions with environmental obstacles.
    \item  \textbf{OSR}: Oracle Success Rate. OSR measures whether the UAV reaches a \(20\text{ m}\) radius of the target along the trajectory, even if it does not stop at the final destination.
    \item  \textbf{NE}: Navigation Error. NE measures the distance between the stop location to the destination.
\end{itemize}
We take the SST and SR as the main metric for DuAl-VLN.

\subsection{Main Comparisons}
\paragraph{Comparison Baselines.}
To validate the effectiveness of our proposed algorithm, we establish both single-UAV and multi-UAV baseline methods. We compare our method with the following four baseline approaches:  
\begin{itemize}[leftmargin=1em]
    \item \textbf{Random}. The UAV randomly selects an action from four possible directions: forward, left, right, up, or down.
    \item \textbf{Cross-Modal Attention (CMA) model}. A Navigation model proposed in AerialVLN~\cite{liu2023aerialvln}, which employs a bi-directional LSTM to jointly process visual inputs and instruction comprehension, predicting the next five waypoints for navigation.
    \item \textbf{TravelUAV}~\cite{wang2024realisticuavvisionlanguagenavigation}. An LLM-based UAV navigation model introduced by ~\cite{wang2024realisticuavvisionlanguagenavigation}. In TravelUAV, the LLM predicts a long-term waypoint, while an LSTM model fills in the intermediate waypoints.
    \item \textbf{TravelUAV (L1 assistant)}~\cite{wang2024realisticuavvisionlanguagenavigation}. A variant of TravelUAV enhanced with the L1-level assistant that provides oracle guidance. At each step, the assistant helps to predict the next action by comparing the UAV’s position and orientation with the ground-truth trajectory, ensuring the UAV stays on the correct path.
\end{itemize}

As shown in Table~\ref{tab:main_comparison}, target-oriented Vision-Language Navigation (VLN) under real-flight conditions in OpenUAV remains a highly challenging task. Existing single-UAV methods, such as CMA~\cite{liu2023aerialvln} and TravelUAV~\cite{wang2024realisticuavvisionlanguagenavigation}, achieve only marginal success and frequently fail to reach the target region. This highlights their limitations in spatial scene understanding and real-world obstacle avoidance. To further assess the performance upper bound, we include TravelUAV (L1 Assistant), a better‑performing variant augmented with oracle-level guidance based on ground-truth trajectories. While it offers improved navigation performance, it still falls short in generalizing to long-horizon planning under natural language instructions.

In contrast, our AeroDuo achieves significantly higher success rates and SST across all evaluation splits. It reaches 16.57\% SR and 14.63\% SST on unseen maps, and 14.86\% SR and 13.54\% SST on unseen objects, demonstrating the advantage of dual-altitude collaboration in complex real-world environments, marking a solid step toward practical UAV-VLN systems using only target description instructions.

\begin{figure}[!t]
    \centering \includegraphics[width=1\linewidth]{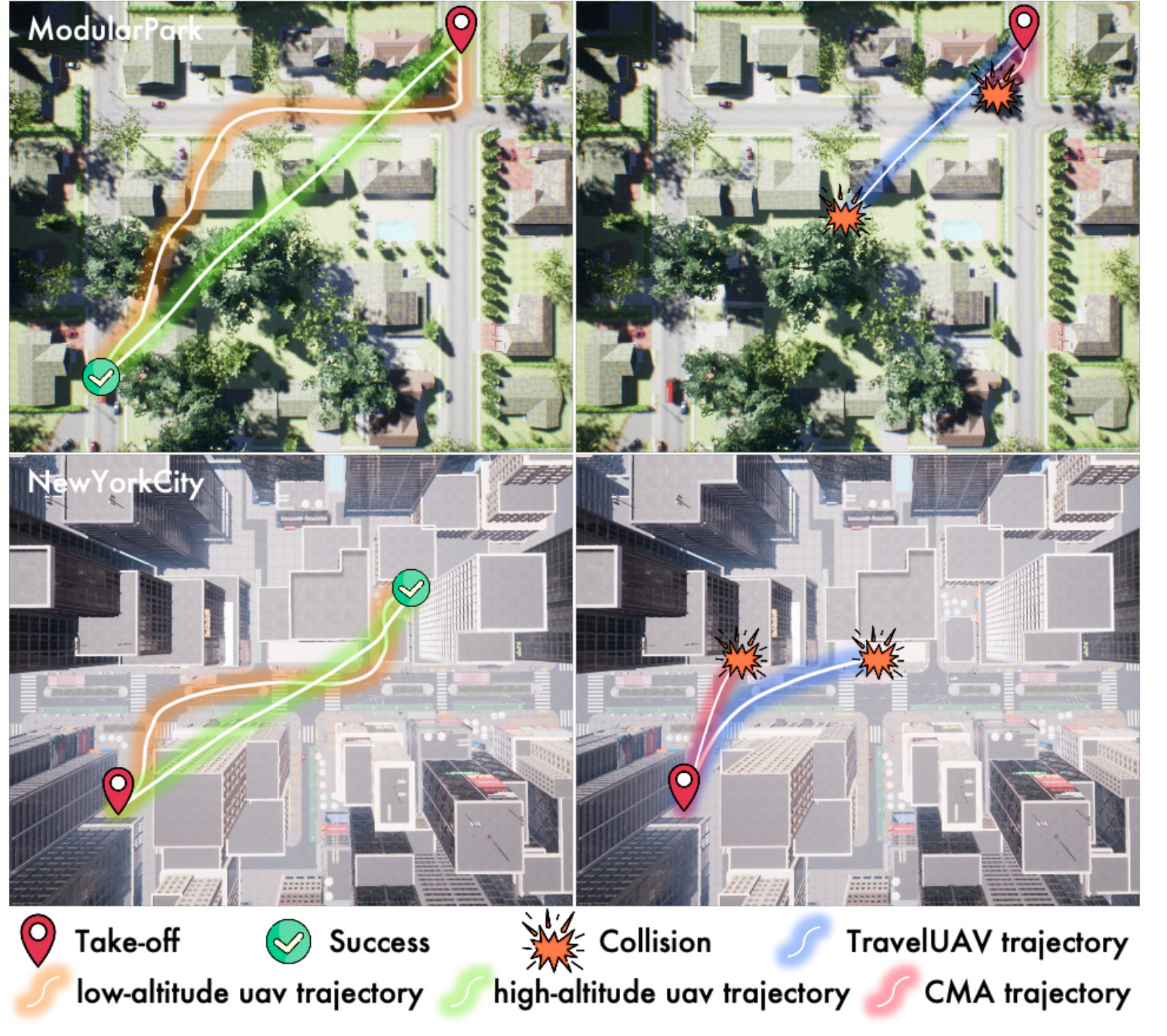}
    \vspace{-4mm}
    \caption{Comparison of UAV target search performance. All methods start from the same take-off position and search for the same target. Left: Our method completes the search without any collisions. Right: TravelUAV and CMA methods result in collisions during the search process.}
\vspace{-4mm}\label{fig:trajectory_comparison}
\end{figure}

\subsection{Ablation Study}
%Macro-Micro UAV collabrative framework.
%Large Language Model-driven Global Map Predictor
We conduct an ablation study to evaluate the contribution of each individual technique. The results are summarized in Table~\ref{tab:ablation_study}, where the following four techniques are examined: MLLM pretraining, global map construction (GMC), key waypoint decision (KWD), and collision-free navigation. In the first rows, the Pilot-LLM without GMC employs the original video encoder~\cite{Qwen2-VL} to observe navigation history. The target distribution is predicted on the current BEV image of $U_h$. The baseline result indicates that relying solely on the CFN leads to poor performance. Comparatively, leveraging the MLLM pretraining and GMC significantly improves the prediction quality of the MLLM, resulting in an 11.41\% increase in SST. In terms of the post-processing and execution of the predicted target point, the results show that the KWD and CFN are critical for the reliable navigation and successful target localization.

\subsection{Qualitative Analysis}

\paragraph{Trajectory Visualization.}
As shown in Fig.~\ref{fig:trajectory_comparison}, compared to single-UAV baseline methods, CMA~\cite{liu2023aerialvln} and TravelUAV~\cite{wang2024realisticuavvisionlanguagenavigation}, our method can accurately locate the target region and plan the flight motion over a relatively long horizon. The two baseline methods both suffer from the collision, failing in most cases.
\begin{figure}[!t]
    \centering \includegraphics[width=1\linewidth]{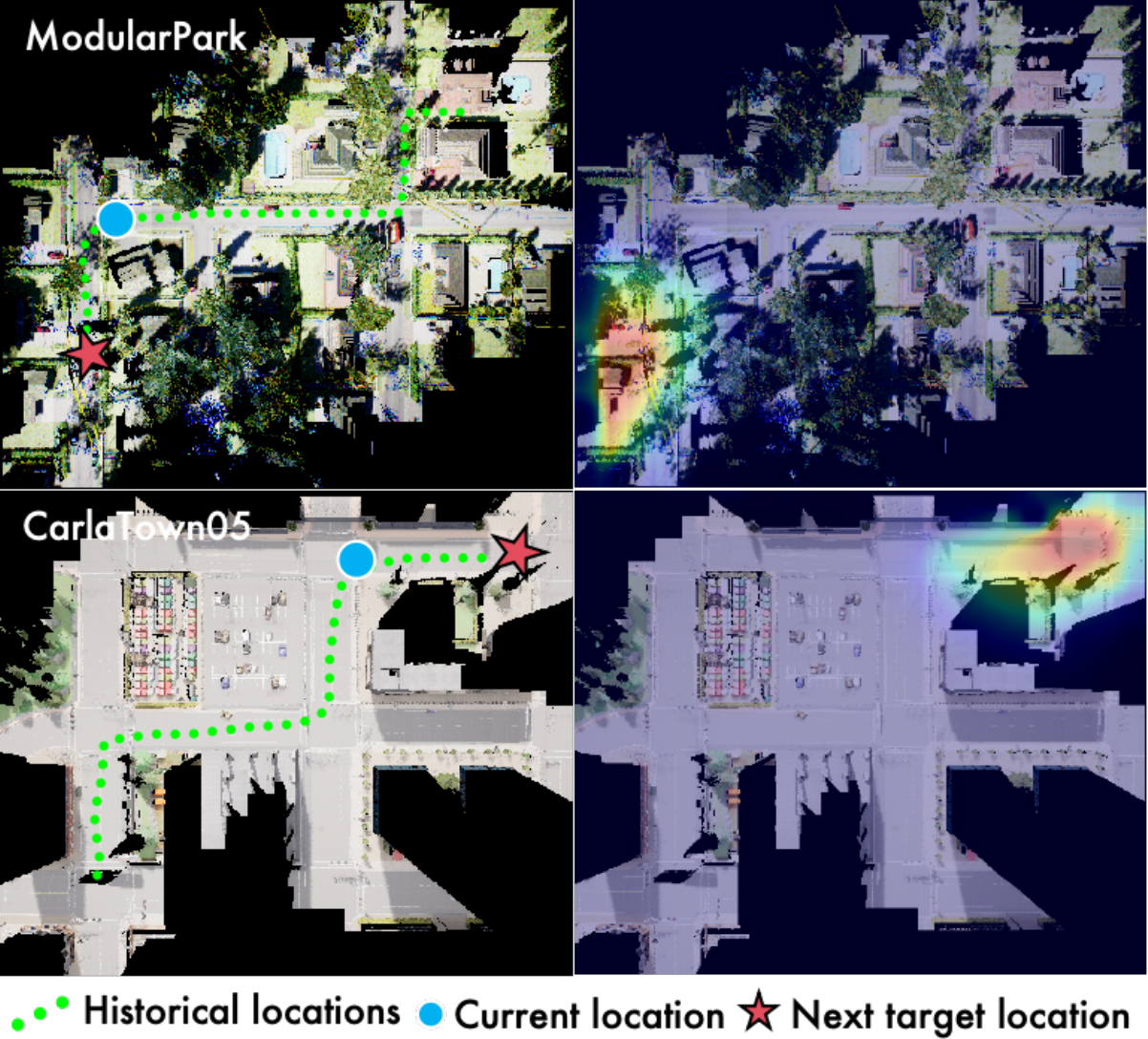}
    \vspace{-4mm}
    \caption{Examples of navigation scenarios and predicted target regions. Left: Orthographic maps and the trajectories of UAV. Right: Predicted heatmaps by Pilot-LLM.}
    \vspace{-4mm}
    \label{fig:prob_map}
\end{figure}

\vspace{1mm}
\paragraph{Target Probability Map Prediction.} Fig.~\ref{fig:prob_map} shows the predicted probability map, where our AeroDuo effectively identifies the destination based on the provided instruction. This highlights the strong capacity of our method to interpret environments from the orthographic map and align observations with complex instructions.

\section{Conclusion}

In this paper, we propose DuAl-VLN, a dual-altitude UAV collaboration task designed to address vision-language navigation challenges in aerial environments. To support this task, we collected the HaL-13k dataset, containing 13,838 synchronized high-low-altitude trajectories, enabling research on altitude-dependent perception and coordination. We present a novel framework, AeroDuo, to handle the DuAl-VLN task. It integrates a high-altitude Pilot-LLM for semantic mapping and a low-altitude agent for obstacle-aware navigation. Experiments show significantly superior success rates over single-UAV baselines, validating collaborative advantages. These findings validate the effectiveness of dual-altitude collaboration and offer a promising direction for aerial embodied AI systems. Future efforts will focus on optimizing the execution efficiency and scalability.

\bibliographystyle{ACM-Reference-Format}
\bibliography{main}

\end{document}